# Trading Confidence: Comprehensive Uncertainty Estimation in Algorithmic Trading

*Completed Research Paper*

**Li Lin**
Nanyang Technological University
50 Nanyang Avenue, Singapore
lli032@e.ntu.edu.sg

**Wang Li Rong**
Nanyang Technological University
Centre for Frontier AI Research, A*STAR
50 Nanyang Avenue, Singapore
lirong002@e.ntu.edu.sg

**Hsuan Fu**
Laval University
Quebec, Canada
hsuan.fu@fsa.ulaval.ca

**Fan Xiuyi**
Nanyang Technological University
50 Nanyang Avenue, Singapore
xyfan@ntu.edu.sg

## Abstract

*Reinforcement Learning (RL) has emerged as a powerful approach in financial trading, enabling agents to learn optimal strategies through direct market interaction. However, financial markets are highly uncertain, with price fluctuations driven by stochastic volatility, model limitations, and regime shifts. Traditional RL models struggle in dynamic environments, often failing to adapt to sudden market disruptions, leading to suboptimal trading decisions. To address this challenge, we propose an uncertainty-aware RL framework that integrates distributional, epistemic, and aleatoric uncertainty estimations. Our approach enhances uncertainty estimation using SHAP-weighted reconstruction uncertainty, MC Dropout, and an LSTM-based technical indicator consensus mechanism. Experimental results on five major U.S. stock indices demonstrate that RL agents equipped with uncertainty estimation significantly outperform traditional models in return and risk management. This study advances uncertainty estimation in RL-based financial trading, with future research extending its application to other asset classes and alternative RL architectures for greater adaptability.*

**Keywords:** Algorithmic Trading, Uncertainty Estimation, Reinforcement Learning

## Introduction

Algorithmic trading has revolutionized financial markets by enabling automated systems to analyse vast amount of data and execute trades with minimal human intervention. The algorithmic trading sector, currently valued at USD 21.06 billion and is expanding at an annual rate of 12.9% (Grand View Research, n.d.). However, this growing reliance on automation introduces new risks, as demonstrated by the 2012 Knight Capital disaster, where a trading algorithm malfunctioned and caused the company to lose USD 440 million in just 45 minutes (Harford, 2012). Though it was initially a technical deployment mistake, this underscored a deeper issue: even advanced trading systems can experience catastrophic failures when they struggle to accurately evaluate and react to various forms of uncertainties. Addressing this challenge requires more robust uncertainty estimation techniques to improve risk-aware decision-making.

Recent developments in machine learning for AI in Finance have led to growing use of Reinforcement Learning (RL) in trading (Zhang et al., 2024, Cartea et al., 2021). The attraction of RL is its capability to acquire good trading strategies via direct engagement with the market and ongoing adjustment. In other words, RL agents learn to trade without explicitly predicting the price or the return of the financial assets they trade, as such predictions are difficult to model accurately in complex market environment. Nonetheless, while RL agents can learn to navigate complex dynamic market environments, their effectiveness heavily depends on their ability to understand and respond to uncertain market behaviours.

Indeed, uncertainty plays an important role in algorithmic trading agent design, as RL agents must function in ever-shifting environments, where historical trends may not consistently forecast future outcomes (Lien et al., 2022). For example, a trading agent trained in a bull market may struggle when faced with a sudden bearish shift, leading to misestimations and significant losses. Similarly, during market crises or geopolitical evens, trading agents may encounter unfamiliar scenarios where their limited understanding increases the risk of poor decisions. These challenges highlight the need for robust uncertainty handling.

In AI in Finance, three fundamental types of uncertainty are particularly relevant:

- First, Aleatoric Uncertainty (Data Uncertainty) represents the inherent randomness. It is irreducible, persisting even with perfect knowledge of the system. Similar to the unpredictability in a coin flip, market movements contain natural stochasticity that cannot be eliminated even with additional data collection (Hüllermeier & Waegeman, 2021).
- Second, Epistemic Uncertainty (Model Uncertainty) stems from the model's limited knowledge of the environment. Unlike aleatoric uncertainty, it can be reduced through additional data or improved modelling approaches (Hüllermeier & Waegeman, 2021). It represents the trading agent's uncertainty about its own predictions and knowledge gaps about market dynamics.
- Third, Distributional Uncertainty occurs when underlying distribution of market data changes over time, particularly in non-stationary environments (Peng & Yang, 2021; Schmitt et al., 2013) This uncertainty poses significant challenges to trading agents as it reflects shifts in market regime or trends (Chen & Tsang, 2020; Cappuccio et al., 2022).

Our research extends the Uncertainty-Augmented Agent Model (Loh et al., 2024), which integrates Reconstruction Uncertainty Estimate (RUE) (Wang et al., 2024, Korte et al., 2024) with Monte Carlo (MC) Dropout (Gal & Ghahramani, 2016) for uncertainty estimation in RL trading agents. While preserving the effective uncertainty estimation using RUE and MC Dropout, we introduce several advancements to develop a more robust uncertainty estimation framework, resulting in significantly improved performance.

1. We enhance the RUE's distributional uncertainty estimation by integrating SHAP (Shapley Additive exPlanations) (Lundberg & Lee, 2017) values. This novel combination allows our model to not only detect distribution shifts but also understand their importance through three distinct combination methods. By linking reconstruction errors with feature importance, our approach enables trading agents to focus on uncertainties arising from truly significant market changes.
2. We introduce a new technical indicator based consensus mechanism for aleatoric uncertainty estimation. This mechanism employs a LSTM architecture that generates forward-looking price predictions, analyses trends through four complementary technical indicators, and aggregates signals through a majority voting system to identify inherent market randomness. This addition completes our uncertainty framework by capturing the market's inherent stochasticity.

Importantly, the entire uncertainty framework functions as a surrogate model for the RL agent. Instead of directly telling the RL trading agent when to buy or sell, the surrogate model helps the agent assess market conditions by providing insights into potential risks and market shifts represented as aleatoric, epistemic and distributional uncertainties. The additional information enables the agent to make more informed, risk-aware decisions while navigating uncertainty without the model itself dictating trade actions.

Building on the Proximal Policy Optimization (PPO) based architecture of the original work (Loh et al., 2024), our enhanced framework as shown in Figure 1 provides trading agents with a more comprehensive understanding of uncertainties. This uncertainty-aware approach is designed to enhance trading performance across diverse market conditions, with emphasis on navigating extreme market events.

To validate our framework, we significantly expand the evaluation scope beyond (Loh et al., 2024) to include five major U.S. market indices: S&P 500 as a large-cap benchmark, NASDAQ Composite with its technology sector focus, Dow Jones Industrial Average representing established companies, Russell 2000 covering the small-cap segment, and Wilshire 5000 providing the broadest market coverage. We test our approach across four major recession periods (1990-1991, 2001, 2007-2009, and 2020), enabling rigorous assessment of our uncertainty estimations during different market conditions and crisis scenarios.

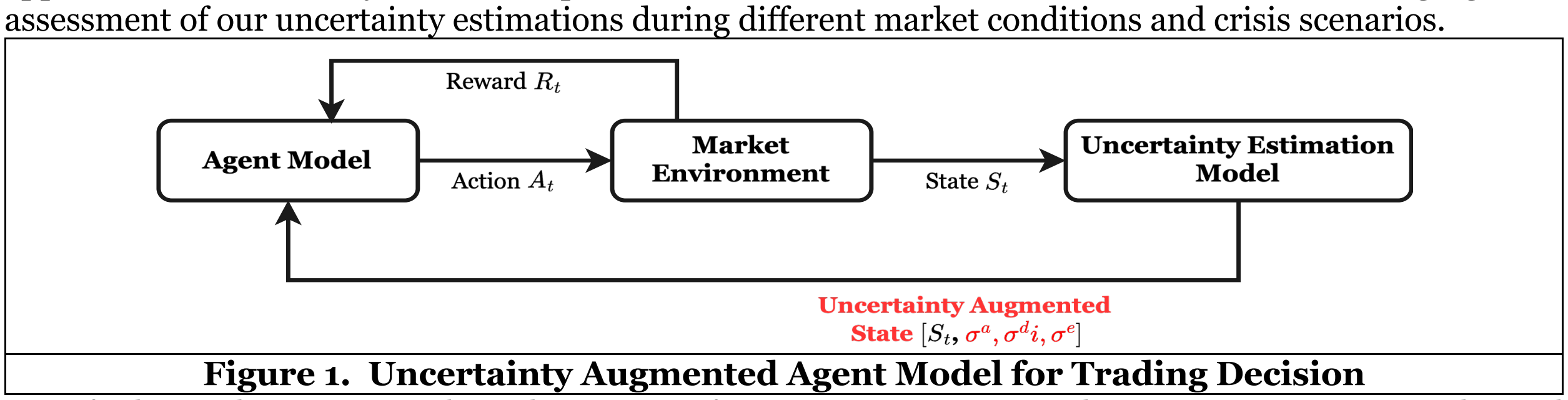

**Figure 1. Uncertainty Augmented Agent Model for Trading Decision**

Our findings demonstrate the advantages of uncertainty-aware trading strategies over traditional approaches. Among the baseline models, the standard PPO RL trading agent achieved only 3.23% average cumulative returns with a mean Sharpe ratio of 0.23, while the ARIMA (AutoRegressive Integrated Moving Average) (Box & Jenkins, 1970) Rule-Based agent suffered significant losses, with negative 22.08% returns, a Sharpe ratio of -0.19, and catastrophic drawdowns of -36.31%. In contrast, an RL agent equipped with our Triple Uncertainty framework, which integrates multiple uncertainty estimation techniques, delivered 21.67% average cumulative returns with a mean Sharpe ratio of 1.50, representing a 6.7x enhancement over PPO and an outperformance against ARIMA Rule-Based.

Our models show remarkable adaptability, particularly during financial crises where traditional methods struggled. During the 2007-2009 financial crisis, the Triple Uncertainty model achieved 30.93% returns, greatly exceeding PPO's 4.28% and ARIMA's -33.75%. Even during the extreme volatility of the COVID-19 recession, our uncertainty-aware models maintained positive returns. This proves their ability to adapt to rapidly changing market conditions. Our results confirm the effectiveness of our framework in navigating complex and unpredictable financial environments.

## Background and Related Work

### *Uncertainty Quantification in Financial Decision-Making*

Recent advances in financial machine learning have increasingly focused on uncertainty quantification as a means to enhance model robustness and decision-making. We structure our discussion of the literature around three primary forms of uncertainty—distributional, epistemic, and aleatoric—each corresponding to distinct theoretical foundations in finance. This mapping strengthens the conceptual bridge between financial and machine learning perspectives, addressing the need for a more rigorous theoretical basis.

Distributional uncertainty captures changes in the underlying data-generating process, such as market regime shifts or structural breaks, and relates closely to regime-dependent asset pricing theories and the Adaptive Market Hypothesis. Markov regime-switching models are a classical approach to this problem, enabling dynamic changes in parameters like volatility and return by modelling latent market states and transition probabilities (Consigli et al., 2017). Alternatively, diffusion models augmented with point processes introduce jump dynamics during periods of instability, accounting for sudden price movements without relying on predefined regimes (Consigli et al., 2017). More recent work in spillover network analysis leverages machine learning, particularly random forests, to detect early warnings of such shifts by modelling interdependencies across markets (An et al., 2025). PCA-based clustering further extends this perspective by identifying emergent regimes from macroeconomic data, offering a data-driven alternative to traditional statistical classification (Akioyamen et al., 2021).

Epistemic uncertainty, which reflects uncertainty in model parameters and structure, parallels financial concerns around forecast reliability and policy sensitivity. Bayesian hierarchical models address this by using posterior distributions derived via MCMC to capture parameter uncertainty and allow inference under multiple economic scenarios. In high-dimensional or sparse data settings, techniques such as Monte Carlo dropout have also proven effective in cryptocurrency markets—by providing uncertainty-aware classification scores that serve as confidence proxies in real-time anomaly detection (Alarab et al., 2021).

Aleatoric uncertainty pertains to the irreducible noise in financial systems, often attributed to microstructure effects or exogenous shocks, and is consistent with the random walk hypothesis. To mitigate its impact, confidence-based filtering techniques have been proposed, which exclude historically similar yet unpredictably divergent samples from training, thereby enhancing predictive precision under high-confidence conditions (Lien et al., 2023). Deep evidential regression provides another promising avenue by jointly modelling predictions and their associated uncertainties, improving transparency in tasks such as loss given default estimation (Nagl et al., 2022). Additionally, Markov decision processes offer a principled way to integrate aleatoric uncertainty into sequential decision-making, aligning with risk-sensitive portfolio management frameworks (Badings et al., 2023)

### *Uncertainty Estimation in Deep Learning*

Monte Carlo (MC) Dropout provides a computationally efficient approach for quantifying epistemic uncertainty in deep neural networks. Gal and Ghahramani (2016) established its theoretical foundation by demonstrating that applying dropout during inference approximates Bayesian inference in deep Gaussian processes. By keeping dropout active during testing and performing multiple forward passes, the variance in predictions quantifies the model's epistemic uncertainty. In financial applications, MC Dropout has proven valuable for decision-making under volatility. Hassan (2024) implemented MC Dropout with Bayesian LSTM models to predict Bitcoin prices while providing confidence intervals. Huang et al. (2024) combined MC Dropout with reinforcement learning through dropout-based probabilistic ensembles with trajectory sampling (DPETS), addressing prediction stability challenges relevant to algorithmic trading.

Reconstruction Uncertainty Estimation (RUE), introduced by Wang et al. (2023), uses an autoencoder-inspired architecture to estimate distributional uncertainty. When applied to algorithmic trading, RUE effectively detects distribution shifts caused by regime changes and macroeconomic events. Compared to ensemble methods, RUE offers computational efficiency for real-time decision-making in financial applications. Beyond finance, Korte et al. (2024) validated RUE's effectiveness in medical imaging, showing the correlation between uncertainty scores and prediction accuracy across domains. However, research gaps remain, such as the limited understanding of feature-specific contributions to uncertainty and the relationship between RUE scores and specific market regimes.

### *Reinforcement Learning for Financial Trading and Consensus from Models*

Reinforcement Learning (RL) agents have shown promising results in financial trading and asset management. Jiang et al. (2024) demonstrated that Q-learning agent - a standard RL technique - outperforms traditional models in asset allocation, achieving superior results in cumulative return, Sharpe ratio, and maximum drawdown. A critical challenge in applying RL to trading is managing uncertainty. Shi et al. (2024) focus on the "sim-to-real gap" through distributionally robust Markov games, while Zhang et al. (2024) identify "distribution shift" as a primary obstacle in offline RL agents for trading.

Proximal Policy Optimization (PPO), introduced by Schulman et al. (2017), is a widely used reinforcement learning algorithm known for its stability and efficiency in complex decision-making tasks, especially in AI in Finance applications. PPO belongs to the policy gradient family of algorithms, unlike value-based methods such as Deep Q-Networks (DQN), but offers better sample efficiency and stability compared to Advantage Actor-Critic (A2C). Recent research (Huang et al., 2022) has suggested that A2C can be viewed as a special case of PPO. PPO's clipped objective function helps prevent destructive policy updates, which is particularly valuable in financial markets where extreme policy shifts could lead to catastrophic trading decisions. While DQN excels at quick learning and adaptation, the exploratory strengths of PPO make it better suited for the complex decision-making required in dynamic trading environments (De La Fuente &

Vidal Guerra, 2024). By refining policy updates, PPO enhances learning reliability, making it particularly well-suited for financial trading where its ability to balance adaptability and risk management has contributed to its adoption in trading agents navigating volatile market conditions.

The concept of consensus in financial forecasting has been explored by Zarnowitz and Lambros (1987), who examined how predictions from different sources can be combined to reduce uncertainty. Gabor (2021) further discusses how consensus plays a role in global finance by aligning institutional investor interests, thereby reducing risks and uncertainties in markets. Both studies demonstrate that consensus serves as a powerful tool for uncertainty reduction in economic variables and market trends.

## Mapping Uncertainties between Finance and AI

Uncertainty is a fundamental challenge in finance as it affects asset pricing, risk management, and decision-making processes. As Lo and Zhang (2017) argued in the adaptive markets hypothesis, financial markets exist in a tension between the theoretical assumption of rationality and documented behavioural anomalies, creating an environment where uncertainty cannot be eliminated through better models alone. The unpredictable nature of investor behaviour, macroeconomic shifts, and external shocks such as geopolitical events further compound uncertainty, making it a persistent challenge in financial systems.

The financial domain has developed sophisticated frameworks for understanding and quantifying different types of uncertainty, which can provide valuable insights when applied to the growing field of AI-based financial technologies. Just as financial analysts assess risk factors across multiple dimensions, RL trading agents must account for different sources of uncertainty to enhance decision-making robustness. We believe “finance uncertainty” can be well mapped to “deep learning uncertainty” as follows.

- ***Aleatoric (Data) Uncertainty*:** Empirical evidence demonstrates that investor behaviour is sensitive to uncertainty attributed to chance or stochastic processes as a distinct dimension, with investors varying in how they perceive this type of randomness (Walters et al., 2023). In machine learning, aleatoric uncertainty acknowledges the intrinsic randomness in the data-generating process (Hofman et al., 2024; Hüllermeier & Waegeman, 2021). The theoretical correspondence emerges from both domains recognizing inherent randomness that persists regardless of data quantity or model sophistication. Just as market prices contain natural stochasticity which cannot be eliminated through additional data collection, aleatoric uncertainty in AI represents the irreducible randomness inherent in data that stems from the stochastic nature of the underlying data-generating process. This mapping suggests that aleatoric uncertainty can serve as a machine learning framework for understanding and modelling the stochastic elements that investors perceive in financial markets.
- ***Epistemic (Model) Uncertainty*:** The mapping between epistemic uncertainty in AI and financial model uncertainty is grounded in established theories of knowledge limitations and model risk. Model uncertainty arises when financial models simplify complex market dynamics through assumptions that may fail under extreme conditions (Hansen & Sargent, 2008), as demonstrated during the 2008 financial crisis. This theoretical framework directly corresponds to epistemic uncertainty in machine learning, where uncertainty arises from limitations in our knowledge about the true underlying model. Parameter uncertainty occurs when estimating model parameters from limited, noisy historical data, often leading to suboptimal portfolio allocations (Pástor & Stambaugh, 2000). Knightian uncertainty represents truly unmeasurable risk where probabilities cannot be assigned due to lack of precedent (Taleb, 2007). The theoretical correspondence lies in the reducible nature of these uncertainties: these financial uncertainties all involve knowledge limitations that can be addressed through improved modelling techniques and additional data, mirroring the reducible nature of epistemic uncertainty in AI systems. Empirical evidence confirms that investors perceive and respond to missing knowledge, skill, or information as a distinct uncertainty dimension (Hirshleifer & Lim, 2023).
- ***Distributional (Query) Uncertainty*:** The theoretical foundation for this mapping emerges from regime change theory in finance and distributional shift theory in machine learning, both addressing the fundamental challenge that mathematical models represent simplified approximations of complex underlying processes (Sullivan, 2015). Regime change uncertainty in finance describes when structural shifts alter market relationships, referring to the tendency of financial markets to "change their behaviour abruptly," with this new behaviour often persisting "for several periods after such a change," frequently corresponding to "different periods in regulation, policy, and other secular changes" (Ang &

Timmermann, 2012). This causes models trained on past data to become ineffective (Nieh, Wu, & Zeng, 2010). This theoretical framework directly corresponds to distributional uncertainty in AI, where in probabilistic modelling, there is always a fundamental challenge that the exact distributions needed to perfectly model a system are rarely available in practice, as the models we use typically represent simplified approximations of more complex real distributions, or we face significant uncertainty about the actual underlying distributions (Coretto & Genest, 2014). Both domains confront the same fundamental theoretical challenge: our mathematical models are simplifications that may not fully capture the complex underlying processes, and both fields require methods that can detect and adapt to fundamental changes in the underlying systems they're modelling. In AI, distributional uncertainty occurs when the statistical properties of the environment diverge from those observed during training, causing learned policies to become suboptimal or harmful. This form of uncertainty presents the greatest challenge in developing robust financial AI systems, as it requires continuous adaptation rather than static modelling approaches.

This mapping provides essential context for our work. By understanding how different types of uncertainty manifest in financial markets, we can develop RL agents that more effectively quantify and respond to them.

# Methodology

We propose a method that estimates three types of uncertainties essential for RL agents, significantly extending (Loh et al., 2024) with two contributions. First, we upgrade RUE's distributional uncertainty estimation through integration with SHAP values. Second, we introduce a novel aleatoric uncertainty estimator based on technical indicator consensus through an LSTM framework.

In our method, the overarching RL agent model is illustrated in Figure 1. Using market information and uncertainty estimations as inputs, the agent determines trading actions. To estimate uncertainties, a separate surrogate model is constructed to forecast market movement (price). However, note that the RL agent does not take these forecasts as direct inputs; only the derived uncertainty estimations are fed into the agent. This design choice prevents the agent from overfitting to specific price predictions, allowing it to focus on risk-aware decision-making rather than unreliable forecasts. The uncertainty estimation model is shown in Figure 2, with three uncertainty estimates highlighted in red.

## *Uncertainty Estimations*

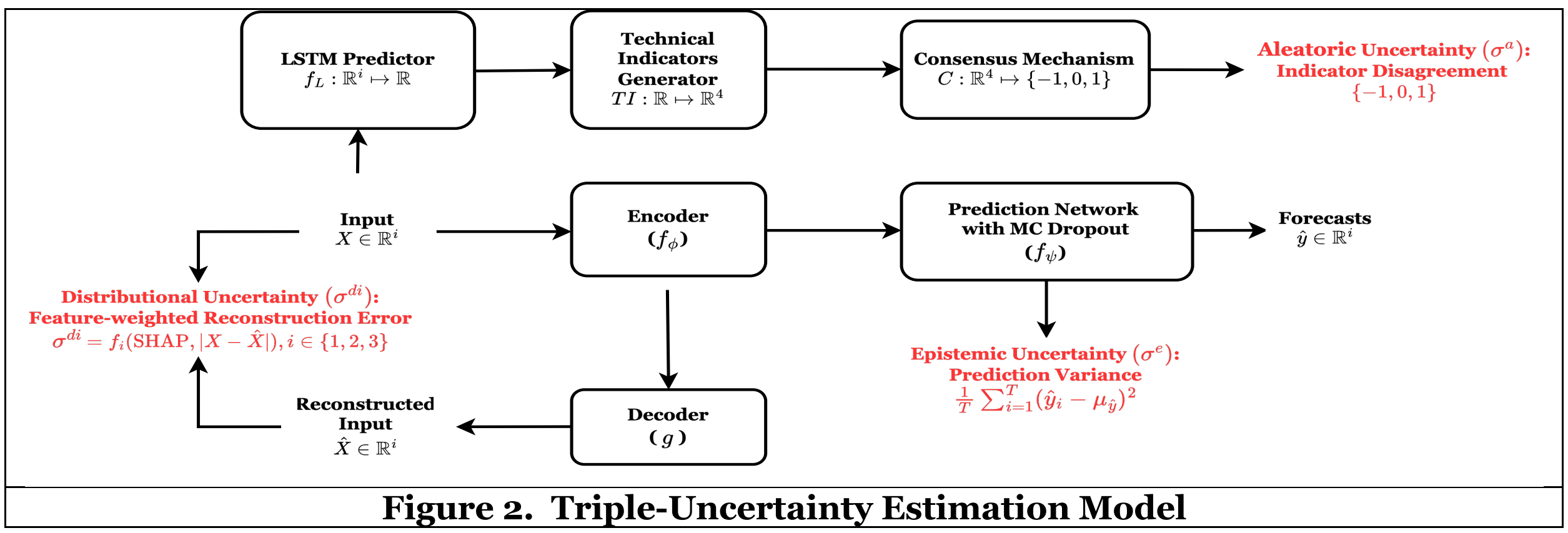

**Figure 2. Triple-Uncertainty Estimation Model**

### Distributional Uncertainty Estimation

Our uncertainty estimation model builds upon RUE architecture. RUE enables early detection of model performance on unfamiliar inputs through an autoencoder-inspired architecture. It employs a multi-layer perceptron (MLP) $f_{\phi,\psi}$ with $N$ layers divided into two components: an encoder $f_\phi$ consists of the first $M$ layers and a decoder $f_\psi$ contains the remaining $N-M$ layers. The model is trained to minimize the difference between actual output $Y$ and predicted output generated from input features $X$:

$$f_{\phi,\psi} = \arg\min_{\phi,\psi} \left|\left|Y - (f_\psi \circ f_\phi)(X)\right|\right|^2,$$

where $f_\phi \circ f_\psi$ denotes the complete prediction model. After the initial training phase, the model rebuilds the input X from its encoded form $f_\phi(X)$. This step aims to reduce the error between the original input and reconstructed output from $g$:

$$g = \arg\min_g \left\|X - \left(g \circ f_\phi\right)(X)\right\|^2 .$$

Models $f_\phi$ and $g$ work together as an autoencoder to rebuild the input data. The method proposes that reconstruction error $\left|x - \left(g \circ f_\phi\right)(x)\right|$ directly scales with prediction error $\left|y - f_{\phi,\psi}(x)\right|$. This relationship makes reconstruction error serve as a distributional uncertainty score:

$$\sigma(x) \coloneqq x - \left(g \circ f_\phi\right)(x).$$

The distributional uncertainty score $|\sigma(x)|$ increases when encountering unfamiliar inputs which reflects the model's confidence level. This approach has two fundamental assumptions:

1. The prediction model $f_{\phi,\psi}$ demonstrates high performance on inputs similar to training data but performs worse on unfamiliar inputs.
2. The reconstruction model $g$ leads to minimal error for familiar inputs but exhibits larger reconstruction errors for unfamiliar data.

Therefore, the reconstruction error $\sigma$ works as a distributional uncertainty indicator. To enhance RUE, we integrate reconstruction error with SHAP values. The idea is that, since $x$ is a vector representing input market data, not all dimensions are equally important for trading strategies. Consequently, failing to reconstruct less important dimensions primarily reflects uncertainties that are less relevant to the task. To ensure that uncertainty estimation focuses on essential information, we use SHAP to identify the most influential features in the price forecasting model $f_{\phi,\psi}$. This approach ensures that reconstruction error $\sigma$ is not treated uniformly but is instead weighted according to SHAP attributions.

We implement three combination methods for computing the final distributional uncertainty $\sigma^{di}$ for our model. Each method combines the reconstruction error $|x - \hat{x}|$ with SHAP values to signal the agent about uncertainty from highly influential features versus less relevant ones. Our first method incorporates multiplication between SHAP values and reconstruction error:

$$\sigma^{d1} = \mathrm{SHAP}(x) \odot \sigma(x)^{\mathrm{T}},$$

where $\mathrm{SHAP}(x)$ represents the feature importance values and $\sigma(x)$ denotes the reconstruction error for each feature inputs. The multiplication ensures that the final uncertainty estimate is amplified only when both components are significant. When only one of them is high, the resulting uncertainty remains proportionally small. This method prevents uncertainty overestimation for less important features.

Our second method normalizes both SHAP values and reconstruction errors to [0,1] before multiplication:

$$\sigma^{d2} = \left(\frac{\mathrm{SHAP}(x) - \min(\mathrm{SHAP})}{\max(\mathrm{SHAP}) - \min(\mathrm{SHAP})}\right) \odot \left(\frac{\sigma(x) - \min(\sigma)}{\max(\sigma) - \min(\sigma)}\right)^{\mathrm{T}}.$$

This method ensures that both SHAP values and reconstruction errors contribute proportionally to the final uncertainty estimate. Without normalization, the agent tends to overreact to features with high SHAP values even when their reconstruction errors are small, or vice versa. By applying min-max scaling, the agent receives more balanced uncertainty signals, which preserve sensitivity to both importance of a feature (SHAP) and how unusual it is (reconstruction error).

Our third method implements a balanced linear combination:

$$\sigma_{d3} = \lambda|\mathrm{SHAP}(x)| + (1-\lambda)|\sigma(x)|,$$

where the weighting factor $\lambda$ controls the contribution of SHAP value and reconstruction. This method prevents uncertainty suppression that can occur in multiplicative methods where one component is small.

The combination of SHAP values and reconstruction error strengthens distributional uncertainty estimation. Higher uncertainty values highlight situations where the model encounters both unfamiliar inputs and significant feature contributions. The trading agent can now focus on truly impactful uncertainties. These filtered insights enable more adaptive and risk-aware decision-making.

### Epistemic Uncertainty Estimation

We follow the integration of MC dropout into RUE introduced in (Loh et al., 2024). This method improves uncertainty awareness for trading agent in scenarios with insufficient or unfamiliar data. It applies dropout during both training and inference phases to approximate Bayesian inference for uncertainty quantification. Dropout layers are integrated where neurons are randomly deactivated with probability $p$. The MC dropout operation at each layer is:

$$z = W \cdot (r \odot h),$$

where $z$ is the output, $w$ denotes the weight matrix, $h$ represents the input features and $r$ is the binary dropout mask with elements set to 1 with probability $p$. During inference, the network performs T forward passes to generate predictions from $\widehat{y_1}, \widehat{y_2}, \ldots, \widehat{y_T}$. The mean is computed as:

$$\mu_{\hat{y}} = \frac{1}{T}\sum_{i=1}^{T} \widehat{y_i} \,.$$

The epistemic uncertainty $\sigma^e$ is then quantified through the variance:

$$Var(\hat{y}) = \frac{1}{T}\sum_{i=1}^{T} \left(\widehat{y_i} - \mu_{\hat{y}}\right)^2,$$

where higher variance indicates greater model uncertainty, particularly when data points deviate from the training distribution.

### Aleatoric Uncertainty Estimation

Our approach to aleatoric uncertainty estimation uses unidirectional LSTM model to predict stock closing prices. For each time $t$, the predictor $f_L$ generates forward-looking closing price forecasts $P_{t+1}, P_{t+2}, P_{t+3}$ that feed into our trend indicator generation process.

Specifically, for a given input sequence of closing price $\{P_{t-n}, \ldots, P_t\}$ , where n is the lookback window. The input sequence $x = (x_{t-n}, \ldots, x_t) \in R^{i \times n}$ consists of feature vectors from the past n days, where each $x_i$ is the feature vector for day $i$. The inputs are processed through two stacked LSTM layers:

- First layer $L_1$ maps input features from $R^i$ to a hidden state in $R^{64}$.
- Second layer $L_2$ further processes the output from $L_1$, mapping it from $R^{64}$ to $R^{32}$, refining the learned features.
- Final dense projection $D$ is a dense layer that maps the output from $L_2$ to a scaler value which is the final close price prediction.

The predictor is trained to minimize the mean squared error:

$$\text{MSE} = \frac{1}{N}\sum \left(P_i - \widehat{P_i}\right)^2.$$

For each timestep t, we generate four technical trend indicators $TI = TI_1, TI_2, TI_3, TI_4$ based on the current price $P_t$ and the predicted prices $\widehat{P_{t+1}}, \widehat{P_{t+2}}, \widehat{P_{t+3}}$. Our framework considers both direct price comparisons and more complex statistical measures to ensure robust trend detection.

The first indicator identifies trend direction through the overall price change. $TI_1$ is end-point change indicator that measures the percentage change between current and final predicted prices:

$$TI_1 = \frac{\widehat{P_{t+3}} - P_t}{P_t} \,.$$

Positive $I_1$ indicates an upward trend , while a negative value suggest downtrend, and zero indicates no significant price movement.

Second, to assess the consistency and quality of price movements, we employ linear regression analysis. This perspective helps to confirm trend direction through consistent price movement through both the slope and goodness of fit. For time stamps t = {0,1,2,3} and corresponding prices P = $\{P_{t,} \widehat{P_{t+1}}, \widehat{P_{t+2}}, \widehat{P_{t+3}}\}$, the linear regression slope $TI_2$ is:

$$TI_2 = \frac{\text{Cov(t, P)}}{\text{Var(t)}},$$

where $\mathrm{Cov}(t, P)$ is the covariance between time and price, $Var(t)$ is the variance of time. The slope $TI_2$ implies an uptrend when positive and a downtrend when negative, with values near zero suggesting neutral movement.
The third indicator $TI_3$ validates trend direction by comparing short-term and long-term moving averages($MA_{\text{short}}$, $MA_{\text{long}}$). This is implemented through a moving average comparison that can identify potential momentum shifts:

$$MA_{\text{short}} = \frac{1}{3}\sum_{i=1}^{3} \widehat{P_{t+\iota}}\ , \qquad MA_{\text{long}} = \frac{1}{4}\sum_{i=0}^{3} \widehat{P_{t+\iota}}\ .$$

To determine the trend, we use the sign function:

$$TI_3 = \mathrm{sign}\left(MA_{\text{short}} - MA_{\text{long}}\right).$$

The sign function returns 1 for uptrend ($MA_{\text{short}} > MA_{\text{long}}$), -1 for downtrend ($MA_{\text{short}} < MA_{\text{long}}$), and 0 for neutral movement.

Finally, to capture trend consistency through daily price movements, we calculate the mean of consecutive price returns:

$$TI_4 = \frac{1}{3}\sum_{i=1}^{3} \frac{P_{t+i} - P_{t+i-1}}{P_{t+i-1}}.$$

A positive $TI_4$ shows consistent daily price increases, while a negative indicates consistent decreases. When daily price changes offset each other or show minimal movement, $TI_4$ is near zero to signal a neutral trend.

After computing the four trend indicators, we define a consensus mechanism $C: R^4 \to \{-1,0,1\}$ to generate the final trend signal. At each time step $t$, the mechanism aggregates the signals from technical indicators $TI_1, TI_2, TI_3, TI_4$ to estimate the trend direction while accounting for uncertainty. The consensus function $C$ is defined as:

$$C(TI_1, TI_2, TI_3, TI_4) = \begin{cases} 1, & if\ at\ least\ three\ indicators\ signal\ an\ uptrend; \\ -1, & if\ at\ least\ three\ indicators\ signal\ a\ downtrend; \\ 0, & otherwise\ (uncertain)\ . \end{cases}$$

This majority voting approach ensures that we only signal a definitive trend when there is a strong agreement among indicators. Whenever fewer than three indicators agree on a direction, we say we are uncertain about the trend.

Since each indicator analyses price trend is from a different perspective (overall trend, movement consistency, price levels and daily patterns), their disagreement signifies that the price movement itself lacks of a consistent pattern. The conflicting signals suggests the presence of inherent randomness in the market environment. The output from this consensus mechanism $C$ directly feeds into our representation of aleatory uncertainty $\sigma^a$, where $\sigma^a \in \{-1, 0, 1\}$ captures the indicator disagreement states.

### *Reinforcement Learning Trading Agent*

In this work, we extend the standard Proximal Policy Optimization (PPO) agent with three uncertainty estimates $\sigma$s. In reinforcement learning, an agent interacts with its environment by receiving a state $s_t$ from the state space $S$. Based on this state, the agent uses its policy $\pi_\theta(a_t|s_t)$ to select an action at from the action space $A$. After taking the action, the agent receives a reward $r_t$ and transitions to a new state $s_{t+1}$. The agent aims to maximize its expected total reward:

$$J(\theta) = E_\pi\left[\sum_{t=0}^{T} \gamma^t r_t\right],$$

where $\gamma$ is the discount factor determines the value of future rewards. To include uncertainty, we expand the state space $S$ by incorporating uncertainty scores generated from our uncertainty estimation model. The agent's model can then be described by its policy function $\pi$:

$$a_t = \pi(s_t)\ ,$$

where $s_t$ represents input state at time $t$ and $a_t$ is the decision action at time $t$.

We add the three uncertainty measures to the state space:

$$a_t = \pi([s_t, \sigma^{di}, \sigma^e, \sigma^a]) ,$$

where:

- $\sigma^{di}$ represents distributional uncertainty with $i \in \{1,2,3\}$ indicating the three variants of distributional uncertainty;
- $\sigma^e$ represents epistemic uncertainty derived from model predictions;
- $\sigma^a$ represents aleatoric uncertainty captured from input features.

The augmented state space $s_t^*$ is now represented as:

$$s_t^* = [s_t, \sigma^d, \sigma^e, \sigma^a].$$

We optimize the policy $\pi_\theta(a_t|s_t)$ using the standard PPO algorithm with the updated state space $s_t^*$. The objective is to maximize the expected total reward while ensuring stable updates to the policy. The clipped surrogate objective for PPO is expressed as:

$$L^{CLIP}(\theta) = E_t\left[\min\left(r_t(\theta)\widehat{A_t}, \text{clip}(r_t(\theta), 1 - \epsilon, 1 + \epsilon)\widehat{A_t}\right)\right],$$

where $r_t(\theta)$ is the probability ratio between the new and old policies:

$$r_t(\theta) = \frac{\pi_\theta(a_t|s_t^*)}{\pi_{\theta_{old}}(a_t|s_t^*)}.$$

and clip is the PPO clipping mechanism that keeps updates gradual, ensuring that the agent adapts steadily. For detailed discussion on PPO agents in general, please refer to Schulman et al. (2017) for details.

## Experiment

### *Dataset*

We target the U.S. stock market due to its different characteristics during recession and non-recession periods to demonstrate the adaptability of our trading agent across varying market conditions. By evaluating performance in both stable and volatile environments, we highlight the agent's ability to manage uncertainty, adjust trading strategies, and maintain robust decision-making under economic fluctuations. Our dataset comprises five major indices: S&P 500, NASDAQ Composite, Dow Jones Industrial Average, Russell 2000 and Wilshire 5000. They cover diverse market segments and capitalization levels for robust evaluation of uncertainty estimation methods:

- **S&P 500** represents large cap stocks and serves as a key benchmark for the U.S. equity market.
- **NASDAQ Composite** is dominated by technology stocks, representing sector-specific volatility.
- **Dow Jones Industrial Average** comprises 30 major and well-established companies and thus provides insights into uncertainty estimation in lower volatility environments.
- **Russell 2000** focuses on small cap stocks and introduces higher volatility and distinct risk patterns.
- **Wilshire 5000** is the broadest U.S stock market index, including both liquid and less liquid securities.

To align with economic cycles, we segment each index's data according to the official recession dates defined by National Bureau of Economic Research (NBER). As shown in Figure 3, the dataset is divided into four overlapping sub-periods centred on major recessions: the 1990-1991 recession, the 2001 dot-com crash, the 2007-2009 financial crisis and the 2020 COVID-19 recession. Each training set includes data from the previous recession to help the model learn from past downturns. The validation set consists the pre-recession period and the initial 30% of the recession period and test set covers the remaining 70%. This segmentation ensures systematic evaluation of uncertainty estimation methods in market crisis. We did not test the Dow Jones, Russel 200 and Wilshire 5000 datasets in Period 1 due to data unavailability.

We retrieve raw historical data from the *yfinance* Python library and compute derived technical indicators. The features used in our trading agent framework were selected to capture various aspects of market dynamics and uncertainty which are crucial to the agent's decision-making and performance. Table 1 presents both the financial indicators and the uncertainty metrics generated by our uncertainty estimation model (denoted with underlining), which are sent to the trading agent. We also provide justifications for the selection of each feature and metric.

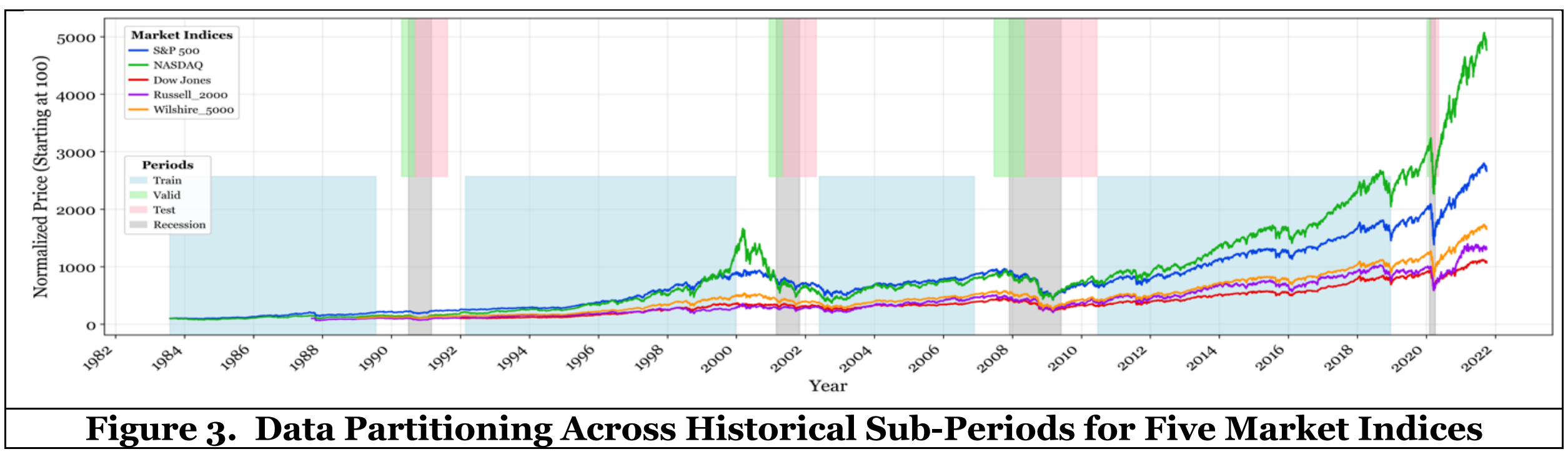

**Figure 3. Data Partitioning Across Historical Sub-Periods for Five Market Indices**

**Table 1. Trading Agent Parameters: Market Features and Uncertainty Metrics**

|  |  |
|---|---|
| Opening Price | The price at which the asset opens during a trading session. It can signal market sentiment and often serves as a key reference for the day's trend. |
| Closing Price | The last traded price of an asset for a given trading session. It is widely used for technical analysis and trading signals. |
| Lowest/Highest Price during the Day | The minimum/maximum price reached by an asset during a trading session. This can indicate market volatility and provide insights into potential support levels. |
| Trading Date & Day of Week | These can influence market behaviour and serve as a feature for temporal analysis. |
| Shares Traded | The volume of shares traded during a session, providing insight into market liquidity and investor activity. High trading volumes may indicate significant market moves. |
| Momentum Indicator (MACD) | This is a trend-following momentum indicator that shows the relationship between two moving averages of an asset's price. |
| MACD Signal Line | The signal line is the moving average of the MACD. When the MACD crosses above this line, it generates a potential buy signal, and vice versa for a sell signal. |
| MACD Histogram | The difference between the MACD line and the signal line, indicating the strength of the trend. A wider histogram suggests a stronger trend, while a smaller histogram signals weakness. |
| Bollinger Upper/Lower Band | Indicates overbought/oversold conditions and helps identify market extremes. |
| 3-Month Treasury Bill Rate | This is a short-term interest rate representing the return on a 3-month U.S. Treasury bond. It reflects broader economic conditions and investor sentiment. |
| 20-Day Moving Average | It captures short-term market trends, smoothing out price movements and providing insights into current market direction. |
| Relative Strength Index (RSI) | A momentum oscillator that measures the speed and change of price movements, indicating whether an asset is overbought or oversold. |
| Average True Range (ATR) | A volatility indicator that measures the average range between the highest and lowest prices over a set period. It helps assess market risk and volatility. |
| RUE (Reconstruction Uncertainty Estimate) | A measure of the uncertainty in predictions, representing the amount of variance between the predicted and actual values, often used to capture model uncertainty. |
| SHAP-RUE (Direct Multiplication, Normalized, Weighted Sum) | It captures the impact of individual market factors on uncertainty, enhancing the agent's understanding of significant features. |
| MC Dropout Prediction Variance | It represents epistemic uncertainty and helps the agent gauge the confidence level of its predictions. |

| Indicator Disagreement | This measures the degree of disagreement between different technical indicators, providing insight into market uncertainty or mixed signals from multiple indicators. |
|---|---|

### Evaluation Metrics

To comprehensively assess the performance of trading agent, we employ three key metrics:

- **Sharpe Ratio (SR)** is calculated using portfolio return $R_p$, risk free rate $R_f$ and standard deviation of portfolio returns $\sigma_p$:
$$SR = \frac{R_p - R_f}{\sigma_p}.$$
This risk-adjusted performance measure accounts for both return and volatility. It considers the excess return per unit of risk. A higher Sharpe Ratio indicates the trading agent is achieving better returns per unit of risk taken.
- **Cumulative Return (SR)** is computed using final portfolio value $V_f$ and initial investment value $V_i$:
$$CR = \frac{V_f - V_i}{V_i}.$$
This metrics measures the total return achieved over the testing period. It reflects the profitability of the agent's strategy. A high cumulative return shows the model's effectiveness in capitalizing on market opportunities.
- **Maximum Drawdown (MDD)** is determined by finding the largest peak to trough decline, where $V_t$ is the portfolio value at time $t$ and $V_p$ is the peak portfolio value observed up to time $t$:
$$MDD = \min\left(\frac{V_t - V_p}{V_p}\right).$$
This represents the largest observed decline from a peak to trough in portfolio value. It highlights the agent's ability to manage downside risk and withstand adverse market strategies.

### Uncertainty Decomposition and Agent Behaviour

To explore the different uncertainties present in market data, we apply our uncertainty decomposition framework across the recession and non-recession periods. This framework distinguishes between distributional uncertainty (enhanced through our RUE integration), epistemic uncertainty (estimated via MC dropout), and aleatoric uncertainty (derived from our novel technical indicator consensus approach). Figure 4 illustrates these uncertainties for the five financial indices. Both distributional uncertainty (green) and epistemic uncertainty (blue) show similar patterns - concentrated and peaked during non-recession periods but becoming wider, more spread out, and shifting rightward during recessions. These are quantified using kernel density estimation plots. Meanwhile, aleatoric uncertainty is uniquely quantified through the Indicator Disagreement Rate (IDR), which represents the percentage of instances where our technical indicators fail to reach consensus, shown as red bars. Our findings show that the IDR nearly doubles during recession periods. This elevation in aleatoric uncertainty indicates higher inherent randomness in market behaviour during downturns. These observations support the importance of incorporating our uncertainty-aware decision-making framework for  agents in volatile environments.

To further demonstrate the practical implications of uncertainty-aware decision-making, we visualize the action values of the P3 agent on the Dow Jones over time in Figure 5. The y-axis represents the agent's action values, while the x-axis shows the corresponding dates. Periods of high estimated uncertainty—identified based on our combined uncertainty metrics—are highlighted in light purple. The chart compares two lines: one representing the uncertainty-aware agent and the other representing the baseline agent. Notably, during high uncertainty periods, the uncertainty-aware agent tends to exhibit lower action values, reflecting a more cautious stance. This behaviour is illustrated by three vertical dashed lines with star markers, where the uncertainty-aware agent's values are significantly lower than those of the baseline. Conversely, during low uncertainty periods (indicated by dashed lines with circular markers), the uncertainty-aware agent often exhibits higher action values, showing greater confidence in decision-making. While this trend is not universally consistent, it illustrates the agent's adaptive behaviour in response to varying levels of market uncertainty. This dynamic adjustment contributes to the performance improvements reported in our results table.

**Figure 4. Uncertainty Decomposition for Recession and Non-Recession Periods**

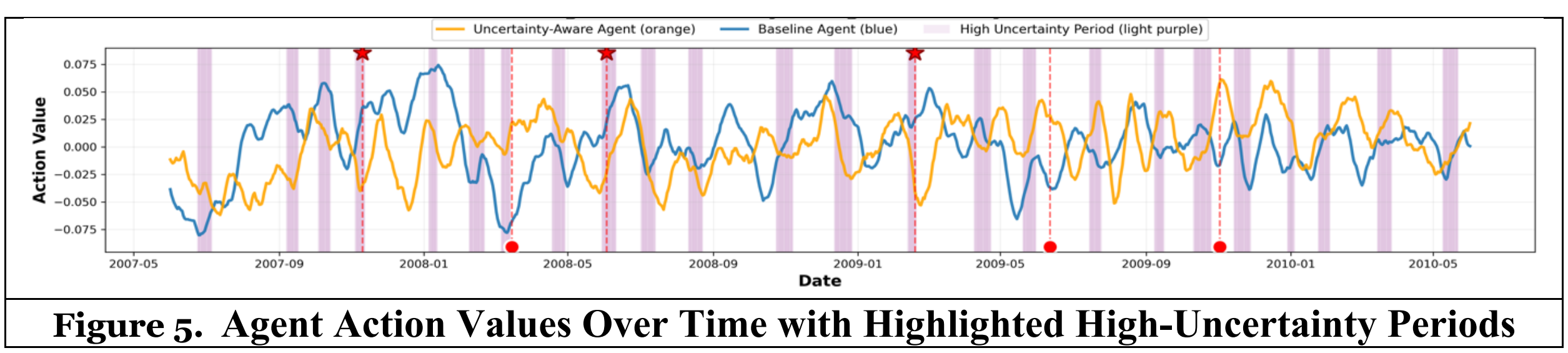

**Figure 5. Agent Action Values Over Time with Highlighted High-Uncertainty Periods**

## *Experiment Design*

To assess the effect of our work and feature importance on the trading agent, we performed thorough experiments across 4 subperiods. Each experiment was repeated five times with different random seeds, and the average of the agent's actions was computed to reduce randomness and ensure reliable outcomes.

- **Method 1: RUE (Reconstruction Uncertainty Estimation):** The state space was augmented with the reconstruction error from the autoencoder, providing a measure of data reliability and market regime shifts.

- **Method 2: MC Dropout:** The state space included prediction variance computed using MC Dropout, capturing model uncertainty in the agent's decision-making process.
- **Method 3: Dual Uncertainty State Space:** The state space is augmented with both RUE and MC Dropout uncertainty estimates as additional features alongside the original state features.
- **Method 4: Direct Multiplication SHAP-RUE (SR):** The state space is augmented with both the SHAP-RUE direct multiplication uncertainty ($\sigma^{d1}$) and MC Dropout uncertainty as additional features alongside the original state features.
- **Method 5: Normalized SHAP-RUE (SR):** The state space includes both the normalized SHAP-RUE uncertainty ($\sigma^{d2}$) and MC Dropout uncertainty as additional features alongside the original state features.
- **Method 6: Weighted Sum SHAP-RUE (SR):** The state space incorporates both the weighted sum SHAP-RUE uncertainty ($\sigma^{d3}$) and MC Dropout uncertainty as additional features alongside the original state features.
- **Method 7: Indicator Disagreement:** The agent's state space is augmented with a measure of disagreement between technical indicators ($\sigma^{a}$), providing market regime uncertainty signals.
- **Method 8: Triple Uncertainty State Space:** The state space is augmented with all three uncertainty measures (RUE, MC Dropout, and indicator disagreement) as additional features alongside the original state features.
- **Method 9: Control (Standard PPO):** The agent operated without any uncertainty estimation, serve as a baseline to assess the effect of uncertainty-aware methods.
- **Method 10: ARIMA with Rule-Based Strategy:** A statistical trading method utilizing an ARIMA model for price forecasting, combined with a straightforward buy-sell strategy that implements a 5-day holding period.

## Results

Results shown in Table 2 demonstrates the substantial advantages of uncertainty-aware trading strategies across four major historical recessions (P1: 1990-1991, P2: 2001 Dot-com Crash, P3: 2007-2009 Financial Crisis, P4: 2020 COVID-19). The best performance for each column is bolded, and the second-best performance is underlined.

The Triple Uncertainty method consistently outperformed the Standard PPO baseline across all periods and indices. In P2 for the S&P 500, it achieved a Sharpe ratio of 1.74 vs. -1.47 for the baseline, while in P3 for the Nasdaq, it delivered 41.44% returns vs. -0.34%. Compared to Dual Uncertainty, Triple Uncertainty further improved results, with higher Sharpe ratios in multiple cases. For example, in P2 for the S&P 500, it reached 1.74 vs. Dual's 0.90, and in P3 for the Nasdaq, it achieved 41.44% vs. 28.47%. However, in some scenarios, such as the Wilshire 5000 in P4, Dual Uncertainty provided better downside protection.

The SHAP-RUE variants—Direct Multiplication, Normalized, and Weighted Sum—consistently outperformed the base RUE model in several scenarios. For instance, in P4 on the S&P 500, Normalized SR achieved a Sharpe ratio of 2.27 versus RUE's 1.95, and in P3 on the Nasdaq, it delivered 41.56% returns versus 28.74%. These results suggest that incorporating explainability features can enhance trading performance. Each SR variant demonstrated unique strengths: Direct Multiplication improved risk control in P2, Normalized SR maximized returns in P3 but with higher risk, and Weighted Sum SR offered a balanced profile in P4. While generally outperforming RUE, some variants underperformed during rapid market recoveries, especially in P4.

To assess the practical benefits of our Triple Uncertainty PPO framework, we compared it to a standard value-based method, Q-learning, using the Dow Jones Period 2 dataset under identical conditions (Fig. 5). PPO achieved a cumulative return of 21.32%, Sharpe ratio of 1.62, and max drawdown of −3.99%, outperforming Q-learning’s 0.85%, 0.12, and −10.34%, respectively. These results underscore the value of integrating uncertainty modelling with policy-gradient methods for financial decision-making.

To further assess the practical deployment viability, we conducted a detailed latency analysis on a single Apple M2 CPU, without GPU acceleration, using high-resolution timers with gradient computations disabled. Average inference times were: 0.289 ms for the autoencoder (distributional and epistemic uncertainty), 0.800 ms for the LSTM price predictor, 0.050 ms for technical indicators (aleatoric uncertainty), and 0.190 ms for the PPO policy. This results in a total end-to-end latency of 1.329 ms per decision, corresponding to a throughput of ~752 decisions per second. While our framework is not designed for ultra-low-latency HFT (1–100 μs), it is highly efficient for real-time and intraday trading on general-purpose CPU-based systems, with no need for specialized hardware or acceleration.

**Table 2. Market Indices Performance Comparison**

| | | S&P 500 | | | | Nasdaq | | | | Dow Jones | | | Russel 2000 | | | Wilshire 5000 | | |
|---|---|---|---|---|---|---|---|---|---|---|---|---|---|---|---|---|---|---|
| | | P1 | P2 | P3 | P4 | P1 | P2 | P3 | P4 | P2 | P3 | P4 | P2 | P3 | P4 | P2 | P3 | P4 |
| Sharpe Ratio | RUE | 1.41 | 0.77 | 0.59 | 1.95 | 1.27 | 1.03 | 0.66 | 2.11 | 1.30 | 0.60 | 1.04 | 1.79 | 0.71 | 1.12 | 0.84 | 0.43 | 1.58 |
| | MC Dropout | 0.94 | 0.46 | 0.56 | 2.05 | 2.18 | 0.75 | 0.71 | 1.43 | 1.67 | 0.60 | 1.05 | **1.92** | 0.69 | 0.59 | 0.91 | 0.57 | 1.62 |
| | Dual Uncertainty | 0.98 | 0.90 | 0.58 | 1.43 | 1.22 | 0.88 | 0.68 | 1.52 | 1.74 | 0.54 | 1.28 | 1.62 | 0.45 | 1.52 | 1.21 | 0.51 | 0.66 |
| | Direct Multiplication SR | 0.88 | 0.62 | 0.44 | 1.76 | 1.16 | 0.65 | 0.58 | 1.85 | 0.58 | 0.55 | 1.10 | 0.98 | 0.41 | 1.50 | 1.08 | 0.47 | 1.57 |
| | Normalized SR | 1.30 | 0.63 | 0.56 | **2.27** | 2.38 | 0.14 | 0.80 | 1.91 | 1.46 | 0.46 | 1.74 | 1.23 | 0.68 | 1.04 | 1.00 | 0.51 | 2.67 |
| | Weighted Sum SR | 1.11 | 0.62 | 0.56 | 2.04 | 2.00 | 1.32 | 0.67 | **1.96** | 1.14 | 0.50 | 1.39 | 0.89 | 0.63 | 1.54 | 1.03 | 0.54 | 0.53 |
| | Indicator Disagreement | 1.53 | 0.67 | 0.64 | 2.01 | 2.22 | 1.47 | 0.51 | 1.48 | 1.69 | 0.57 | 1.29 | 1.51 | **0.76** | 0.88 | 0.69 | 0.60 | **1.79** |
| | Triple Uncertainty | **1.59** | **1.74** | **0.65** | 1.42 | **2.51** | **1.67** | **0.83** | **1.96** | **2.09** | **0.71** | **2.18** | 1.77 | 0.75 | **1.70** | **1.41** | **0.80** | 1.73 |
| | Standard PPO | 0.60 | -1.47 | 0.49 | 0.86 | 2.36 | -0.39 | -0.29 | 0.88 | 0.61 | 0.35 | 0.76 | -1.58 | 0.55 | -1.65 | 0.73 | 0.10 | 0.74 |
| | ARIMA | -0.54 | -0.88 | -0.06 | -0.09 | 0.93 | -1.06 | -0.39 | 0.33 | -0.30 | -0.06 | -0.02 | 0.09 | -0.44 | -0.14 | -0.44 | -0.39 | 0.06 |
| | | S&P 500 | | | | Nasdaq | | | | Dow Jones | | | Russel 2000 | | | Wilshire 5000 | | |
| | | P1 | P2 | P3 | P4 | P1 | P2 | P3 | P4 | P2 | P3 | P4 | P2 | P3 | P4 | P2 | P3 | P4 |
| Percentage of Cumulative Return | RUE | 8.69 | 3.78 | 12.82 | 11.04 | 12.91 | 15.30 | 28.74 | 11.17 | 13.08 | 25.08 | 7.52 | **25.41** | 19.24 | 7.22 | 15.68 | 16.09 | 10.61 |
| | MC Dropout | 5.24 | 3.85 | 22.76 | 8.52 | 13.93 | 9.84 | 27.43 | 12.38 | 21.73 | 18.25 | 7.82 | 23.66 | 29.31 | 0.89 | 19.18 | 18.11 | 11.68 |
| | Dual Uncertainty | 6.91 | 3.67 | 25.80 | 8.79 | 18.93 | 12.35 | 28.47 | 9.51 | 27.27 | 18.61 | 6.36 | 22.45 | 16.24 | **11.26** | 18.43 | 15.61 | 13.51 |
| | Direct Multiplication SR | 2.58 | 1.33 | 6.12 | **11.81** | 8.21 | 1.11 | 19.72 | 11.84 | 19.37 | 18.45 | 5.85 | 6.86 | 7.71 | 7.99 | 17.28 | 16.38 | 10.21 |
| | Normalized SR | 5.54 | 8.56 | 19.71 | 10.96 | **23.94** | 0.95 | **41.56** | **13.32** | 25.69 | 17.52 | 9.53 | 12.63 | 33.13 | 6.08 | 15.28 | 24.30 | **20.99** |
| | Weighted Sum SR | 7.77 | 3.04 | 18.91 | 5.31 | 5.08 | 34.90 | 26.46 | 7.06 | 13.91 | 20.75 | 8.57 | 3.33 | 20.09 | 0.84 | 9.00 | 16.68 | 2.04 |
| | Indicator Disagreement | 6.60 | 9.47 | 30.12 | 9.03 | 14.61 | 38.28 | 19.24 | 9.90 | 29.28 | 24.84 | 9.38 | 16.73 | **41.72** | 0.96 | 6.75 | 17.49 | 12.52 |
| | Triple Uncertainty | **12.33** | **11.04** | **31.12** | 10.38 | 22.35 | **38.96** | 41.44 | 10.48 | **33.73** | **29.48** | **12.63** | 13.52 | 37.57 | 9.92 | **24.05** | **25.02** | 10.18 |
| | Standard PPO | 0.23 | -0.06 | 20.92 | 3.33 | 4.20 | -0.29 | -0.34 | 1.73 | 11.32 | 5.19 | 2.17 | -3.43 | 13.61 | -0.40 | 7.51 | 1.33 | 8.31 |
| | ARIMA | -7.14 | -13.21 | -12.78 | 3.07 | 0.93 | -25.65 | -39.59 | 1.02 | -37.84 | -24.04 | -21.15 | 4.16 | -70.27 | -52.74 | -58.61 | -58.61 | -11.06 |
| | | S&P 500 | | | | Nasdaq | | | | Dow Jones | | | Russel 2000 | | | Wilshire 5000 | | |
| | | P1 | P2 | P3 | P4 | P1 | P2 | P3 | P4 | P2 | P3 | P4 | P2 | P3 | P4 | P2 | P3 | P4 |
| Percentage of Maximum Drawdown | RUE | -4.37 | -3.03 | -12.46 | -9.92 | -3.75 | -7.54 | -24.86 | -5.51 | -4.37 | -20.82 | -11.00 | **-1.43** | -16.18 | -12.44 | -10.70 | -28.94 | -8.34 |
| | MC Dropout | -2.86 | -5.98 | -22.47 | **-3.73** | -6.25 | -7.98 | -18.12 | -12.52 | -6.23 | -15.49 | -11.77 | -3.93 | -19.05 | -1.07 | **-7.61** | -25.18 | -8.20 |
| | Dual Uncertainty | -1.65 | **-0.91** | -16.31 | -9.81 | -6.78 | -1.75 | -22.24 | -10.13 | -5.79 | -18.01 | **-4.04** | -4.78 | -24.09 | -8.35 | -9.21 | **-16.38** | **-6.64** |
| | Direct Multiplication SR | -1.47 | -1.16 | **-7.44** | -10.02 | -6.56 | -2.26 | -21.86 | -6.12 | -14.21 | -14.24 | -6.58 | -6.36 | -19.75 | -4.78 | -9.18 | -21.43 | -8.03 |
| | Normalized SR | -1.89 | -10.92 | -15.68 | -4.07 | -5.86 | -11.05 | -19.98 | -7.18 | -10.78 | -24.21 | -9.00 | -11.81 | -16.43 | -3.81 | -9.18 | -27.98 | -7.08 |
| | Weighted Sum SR | -3.65 | -5.50 | -17.14 | -2.36 | -3.16 | -11.50 | -22.94 | -4.05 | -5.54 | -26.32 | -7.57 | -2.63 | **-15.34** | **-0.73** | -8.21 | -23.53 | -9.51 |
| | Indicator Disagreement | -2.90 | -12.39 | -24.84 | -4.00 | -6.17 | -8.20 | -21.89 | -10.53 | -6.99 | -21.45 | -11.67 | -4.79 | -23.32 | -8.37 | -10.79 | -20.20 | -8.60 |
| | Triple Uncertainty | *-1.77* | -5.21 | -21.31 | -10.02 | -5.51 | -7.80 | -22.46 | -8.77 | **-5.12** | -18.02 | -4.90 | -2.48 | -17.62 | -6.32 | -9.85 | -25.23 | -7.80 |
| | Standard PPO | **-0.20** | -16.53 | -16.34 | -16.17 | **-1.15** | **-0.78** | **-0.60** | **-2.92** | -19.80 | **-2.28** | -4.24 | -4.93 | -17.03 | -6.69 | -12.99 | -35.83 | -13.36 |
| | ARIMA | -11.47 | -1.06 | -26.43 | -5.42 | -6.67 | -31.67 | -52.76 | -17.56 | -56.27 | -60.37 | -46.90 | -29.91 | -81.98 | -61.90 | -65.16 | -69.96 | -40.44 |

# Conclusion

Reinforcement Learning (RL) has demonstrated significant potential in financial trading by allowing agents to autonomously learn and adapt trading strategies. However, handling financial uncertainty remains a critical challenge for RL-based trading agents. Markets exhibit various forms of uncertainty—ranging from inherent randomness in price movements to model limitations and regime shifts—making robust uncertainty estimation essential for effective decision-making.

This work presents an uncertainty-aware RL trading framework that systematically incorporates three key types of uncertainty: distributional, epistemic, and aleatoric uncertainties. By mapping financial uncertainty concepts to their machine learning counterparts, we develop a comprehensive estimation mechanism that enhances the robustness of trading agents. Our method extends prior uncertainty estimation approaches by integrating SHAP-weighted reconstruction uncertainty, MC Dropout for epistemic uncertainty, and an LSTM-based technical indicator consensus for aleatoric uncertainty. These enhancements allow RL agents to navigate volatile market conditions more effectively.

Our extensive empirical evaluations, covering multiple U.S. stock indices and historical recession periods, validate the effectiveness of uncertainty-aware trading strategies. Results show that RL agents equipped with uncertainty estimation outperform traditional models, achieving higher cumulative returns and Sharpe ratios while demonstrating resilience during financial crises. Notably, the Triple Uncertainty framework consistently delivers superior risk-adjusted returns, confirming that incorporating uncertainty estimations into RL trading agents leads to more stable and adaptive decision-making.

Moving forward, future research should explore broader asset classes beyond equities, such as commodities, forex, and cryptocurrencies, to assess the generalizability of uncertainty-aware RL agents. Additionally, investigating alternative agent architectures—including transformer-based RL models and ensemble reinforcement learning—could further enhance adaptability and performance. By continuing to refine uncertainty estimation techniques, we move toward more resilient, risk-aware AI-driven trading strategies that can better navigate the complexities of financial markets.